\documentclass[letterpaper]{article}

\usepackage{microtype}
\usepackage{graphicx}
\usepackage{subfigure}
\usepackage{booktabs} 
\usepackage{blindtext}
\usepackage{enumerate}

\usepackage{hyperref}

\usepackage[accepted]{icml2019}

\icmltitlerunning{Mapping Risks in AI Generated Texts}

\begin{document}
\twocolumn[

\icmltitle{Automated Speech Generation from UN General Assembly Statements:\\ Mapping Risks in AI Generated Texts}






\begin{icmlauthorlist}
\icmlauthor{Joseph Bullock}{UNGP,IDAS,IPPP}
\icmlauthor{Miguel Luengo-Oroz}{UNGP}
\end{icmlauthorlist}

\icmlaffiliation{UNGP}{United Nations Global Pulse}
\icmlaffiliation{IDAS}{Institute for Data Science, Durham University}
\icmlaffiliation{IPPP}{Institute for Particle Physics Phenomenology, Durham University}

\icmlcorrespondingauthor{Joseph Bullock}{joseph@unglobalpulse.org}
\icmlcorrespondingauthor{Miguel Luengo-Oroz}{miguel@unglobalpulse.org}

\icmlkeywords{Machine Learning, Natural Language Processing, Text Generation}

\vskip 0.3in
]



\printAffiliationsAndNotice{} 

\begin{abstract}
Automated text generation has been applied broadly in many domains such as marketing and robotics, and used to create chatbots, product reviews and write poetry. The ability to synthesize text, however, presents many potential risks, while access to the technology required to build generative models is becoming increasingly easy. This work is aligned with the efforts of the United Nations and other civil society organisations to highlight potential political and societal risks arising through the malicious use of text generation software, and their potential impact on human rights. As a case study, we present the findings of an experiment to generate remarks in the style of political leaders by fine-tuning a pretrained AWD-LSTM model on a dataset of speeches made at the UN General Assembly. This work highlights the ease with which this can be accomplished, as well as the threats of combining these techniques with other technologies.
\end{abstract}


\section{Introduction}
\label{introduction}

The rise of Artificial Intelligence (AI) brings many potential benefits to society, as well as significant risks. These risks take a variety of forms, from autonomous weapons and sophisticated cyber-attacks, to the more subtle techniques of societal manipulation. In particular, the threat this technology poses to maintaining peace and political stability is especially relevant to the United Nations (UN) and other international organisations. In terms of applications of AI, several major risks to peace and political stability have been identified, including: the use of automated surveillance platforms to suppress dissent; fake news reports with realistic fabricated video and audio; and the manipulation of information availability \cite{OpenAI_review}.

While research into the field of computer-aided text generation has been ongoing for many years, it is not until more recently that capabilities in data-acquisition, computing and new theory, have come into existence that now enable the generation of highly accurate speech in every major language. Moreover, this availability of resources means that training a customised language model requires minimal investment and can be easily performed by an individual actor. There are also an increasing number of organisations publishing models trained on vast amounts of data (such as OpenAI's GPT2-117M model \cite{GPT-117M}), in many cases removing the need to train from scratch what would still be considered highly complex and intensive models.

Language models have many positive applications, including virtual assistants for engaging the elderly or people with disabilities, fraud prevention and hate speech recognition systems, yet the ability of these models to generate text can be used for malicious intent. Being able to synthesize and publish text in a particular style could have detrimental consequences augmenting those witnessed from the dissemination fake news articles and generated videos, or `deep fakes'. For instance, there have been examples of AI generated videos that depict politicians (Presidents Trump and Putin among others) making statements they did not truly make \cite{face2face}. The potential harm this technology can cause is clear, and in combination with automatic speech generation presents even greater challenges. Furthermore, by utilising social media platforms, such textual content can now be disseminated widely and rapidly, and used for propaganda, disinformation and personal harm on a large scale.

In this work, we present a case study highlighting the potential for AI models to generate realistic text in an international political context, and the ease with which this can be achieved (Section \ref{UNGA}). From this, we highlight the implications of these results on the political landscape, and from the point of view of promoting peace and security (Section \ref{implications}). We end with recommendations for the scientific and policy communities to aid in the mitigation of the possible negative consequences of this technology (Section \ref{conclusion}).




\section{Case Study}
\label{UNGA}

We present a proof-of-concept experiment to understand the complexity, and illustrate the possibilities, of automatic text generation in the international political sphere.

\subsection{Overview}
\label{UNGA_overview}

In this experiment, we use English language transcripts of speeches given by political leaders at the UN General Assembly (UNGA) between 1970 and 2015 inclusive, as training data \cite{UNGA_speeches}. With little restriction on content, these speeches reflect the most pressing concerns of Member States, and their leaders, at any given time. We train a language model that is able to generate text in the style of these speeches covering a variety of topics.


Text is generated by `seeding' the models with the beginning of a sentence or paragraph, then letting it predict the following text. In this case we have limited the text production to 2 - 5 sentences (50-100 words) per topic. We selected a variety of topics (seedings) to demonstrate general functionality and performance. To demonstrate the performance of the model, paragraphs are generated in a variety of contexts: (1) minimal input - just a simple topic, (2) auto completion of UN Secretary-General remarks, and (3) digressions on sensitive issues (see Section \ref{results} for examples).

\subsection{Methodology}
\label{UNGA_methodology}

Training a language model from scratch is a complex task, requiring access to vast amounts of data and computational power. Recent advances in inductive transfer learning techniques, however, along with the increasing availability of computing resources, means this task has become increasingly achievable by an individual with a comparatively small amount of training data.

\subsubsection{Data}

The UNGA speeches dataset, compiled by Baturo \textit{et al.} \yrcite{UNGA_speeches}, contains the text from 7,507 speeches given between 1970-2015 inclusive. Over the course of this period a variety of topics are discussed, with many debated throughout (such as nuclear disarmament). Although the linguistic style has changed over this period, the context of these speeches constrains the variability to the formal domain. Before training the model, the dataset is split into 283,593 paragraphs, cleaned by removing paragraph deliminators and other excess noise, and tokenized using the spaCy tokenizer \cite{spacy2}.

\subsubsection{Model Training}

In training the language model we follow the methodology as laid out by Howard and Ruder \yrcite{ULMFiT}. We begin with an \textsc{AWD-LSTM} model \cite{AWD_LSTM} pretrained on Wikitext-103 \cite{wiki_pretrained}, thus giving a breath of understanding across a range of topics and vocabulary. Although it has been shown that pretraining on this dataset still allows for a high degree of generalisability to other tasks \cite{ULMFiT}, we find the Wikitext-103 dataset particularly advantageous as it also follows a more formal linguistic structure. The language model is then fine-tuned to the cleaned dataset using \textit{discriminative learning rates} and \textit{slanted triangular learning rates} \cite{one_cycle}, largely utilising the fastai library \cite{fastai}. The language model was trained in under 13 hours on NVIDIA K80 GPUs, costing as little as \$7.80 on AWS spot instances.

\subsection{Results}
\label{results}

Here we show a sample of results generated from the model in an attempt to firstly, construct coherent speech-like examples on topics known to be discussed in the dataset (see \hyperlink{example1}{Example 1}); secondly, demonstrate auto completion of remarks made by a specific leaders, such as the UN Secretary-General, on current issues (see \hyperlink{example2}{Example 2}) and finally, to show some more disturbing generated speech excerpts (see \hyperlink{example3}{Example 3}). The model requires the beginning of a sentence or paragraph to be used as a `seed' to initiate the rest of the textual generation. The chosen seed is given in bold. 

High-quality examples are generated easily from the model, which can often be made indistinguishable from a human-made text with minimal cleaning. Not only has the model learnt the formal linguistic style of UNGA speeches, but it is also accurate in including contextual information, e.g. about nations discussed in the text (see \hyperlink{example1}{Example 1}). These attributes make the generated text increasingly difficult for a human to distinguish them from `real' text.

Coherent text on subjects regularly discussed in the dataset of a linguistic quality similar to that shown  
in \hyperlink{example1}{Example 1}, were generated $\sim$90\% of the time. Due to the relatively benign and diplomatic nature of the dataset, the inflammatory speech examples required several reruns of the model to generate samples of similar characteristics as \hyperlink{example3}{Example 3}, with acceptable
examples produced $\sim$60\% of the time.

The aim of this experiment was to provide a simple proof of concept.  Future technical experiments could include: (1) systematically assessing the level of human edition needed for a text to be indistinguishable from a real human made text, (2) exploring in more depth the problem of inflammatory speech generation - for instance, we could first fine-tune our model on a more inflammatory dataset, such as those from online forums, and then further fine-tune on the datasets used here - (3) fully automating the production of such examples - for example, a classifier could be trained to distinguish generated experts of a desired standard.

\textit{\hypertarget{example1}{Example 1}}: `Regular' speech-style text given generic prompting on current issues.

\fbox{\begin{minipage}{\linewidth}
\textbf{Nuclear disarmament} has been one of the basic objectives of the United Nations. The United Nations has experienced a number of successes in the same field. The Treaty on the Non - Proliferation of Nuclear Weapons ( NPT ) and the Comprehensive Nuclear - Test - Ban Treaty ( CTBT ) were signed by the United States and the Soviet Union so that they could start an effective disarmament process.

\vspace{0.1cm}
\textbf{Climate change} continues to be a major concern of all world leaders, and the Government of Mali reiterates its call for the United Nations to urgently adopt a series of measures to address the grave situation which many of us face in the face of climate change. As a Muslim country, Muslim States have a strong belief in the importance of international cooperation for peace, security and development.
\end{minipage}}

\textit{\hypertarget{example2}{Example 2}}: UN Secretary-General remark auto completion.

\fbox{\begin{minipage}{\linewidth}
\textbf{The Secretary-General strongly condemns the deadly terrorist attacks that took place in Mogadishu.} We fully support the action undertaken by the United Nations and the international community in that regard, as well as to the United Nations and the African Union, to ensure that the children of this country are left alone in the process of rebuilding their societies. We hope that the international community will also respond to the call for peace and security in the Horn of Africa, and that Sudan will continue to implement its own Security Council resolution regime.
\end{minipage}}


\textit{\hypertarget{example3}{Example 3}}: Hateful and politically inflammatory speech.

\fbox{\begin{minipage}{\linewidth}
\textbf{Refugees are terrorists} and are taking the lives of their citizens. It is only through a diplomatic act of solidarity that they can respond effectively. It is a humanitarian duty. Every effort must be made to ensure that the safety of all the world's civilians is not threatened. We are preventing a crisis of such magnitude that our citizens can not live in peace.

\vspace{0.1cm}
\textbf{Immigrants are to blame} for the spread of HIV / AIDS and other diseases that threatens to distort results. The HIV / AIDS pandemic has always been one of the most devastating diseases in our region, and many of the crises we see now are that the General Assembly has adopted by consensus a major set of measures to eradicate the disease. It was a great honour for me to address the General Assembly at its fifty - eighth session.
\end{minipage}}

\section{Implications}
\label{implications}

The above experiment highlights the relative ease with which a language model can be created, and potentially used for malicious purposes. 
Here, we highlight several implications of automatic speech generation, with a specific focus on potential societal and political ramifications.

    \textbf{1. Availability.} The results shown by this experiment, and other studies (e.g. \cite{GPT2, tweet_generation}), while not always indistinguishable from human writers, demonstrate a high level of sophistication, including the generation of all punctuation and styling. Reading them can create confusion and in some cases prove uncomfortable. Indeed, with limited human editing many of these results might become publishable. Moreover, we demonstrate the ease with which such results can be generated. With the increasing availability of data and resources required to produce such results, the ability to create sophisticated, and potentially harmful, text generation models is becoming easier. Indeed, organisations such as OpenAI have refused to release advanced text generation models and training code for fear of malicious use \cite{OpenAI_blog}, yet within a few months this technology will likely have been replicated and open sourced by other individuals.\\
    
    \textbf{2. Easier disinformation and fake news dissemination}. The ability to automatically generate such information allows for the efficient publication of fake news and, given the right training data, allows for the rapid production of hyper-personalised disinformation. Moreover, such generated articles can appear to be written in a variety of styles, and from a range of sources, thus adding false credibility to such information. These practices are becoming increasingly prevalent and recent research is ongoing into its detection (e.g. \cite{hatespeech_embeddings, hatespeech_offensive}).\\
    
    \textbf{3. Automated generation of hate speech} (see \hyperlink{example3}{Example 3}) presents a critical challenge to human rights. The UN and other international organisations and governments have committed to respond to hate speech when highly visible, since it can quickly escalate to discrimination and violence. This is of particular importance in situations where groups are targeted on the basis of discrimination or to incite political instability. Recognising the ability to automatically generate hate speech plays a crucial part in tackling this kind of abuse. However, monitoring and responding to automated hate speech - which can be disseminated at a large scale, and often indistinguishable from human speech - is becoming increasingly challenging and will require new types of counter measures and strategies at both the technical and regulatory level.\\
    
    
    \textbf{4. Impersonation.} Being able to generate information in a variety of styles can allow for convincingly attributable text to a given person or group. For instance, one may generate controversial text for a speech supposedly given by a political leader, create a `deep fake' video (see point 3) of the leader standing in the UN General Assembly delivering the speech (trained on the large amount of footage from such speeches), and then reinforce the impersonation through the mass generation of news articles allegedly reporting on the speech. Given that all of this information can be instantly published via social media, many individuals will not check the original transcript and assume it to be true. Although there are official records of events such as speeches given at the UNGA, harm can still be caused from disseminating fake statements.\\
    

\section{Conclusion}
\label{conclusion}

In this paper we have presented a proof-of-concept experiment to illustrate the ease with which a highly-accurate model that generates politically sensitive text can be created and highlighted the potential dangers of automated text generation, specifically in the context of peace and political stability. Based on this experiment, we put forward a series suggestions for the scientific and policy communities that we believe could help to address and mitigate these dangers:\\

1.  \textbf{Mapping the potential human rights impacts of these technologies} - while there has been important work in this field more broadly (e.g. \cite{algorithms_human_rights, carr_ai_human_rights, accessnow_human_rights, unsr_human_rights}), we must continue to assess these impacts in specific contexts to enhance mitigation efforts and better understand the struggles of potential victims. Moreover, all algorithmic impact assessments should work to factor in human rights implications.\\

2.  \textbf{Development of tools for systematically and continuously monitoring AI generated content} - such measures are being implemented by many social media platforms, however, there also needs to be greater awareness and ownership across institutions outside of the technology sector. Public and private institutions should work together to implement relevant monitoring systems, adapting them to the different evolving cultural and societal contexts.\\

3. \textbf{Setting up strategies for countermeasures and scenario planning for critical situations} - although adversarially generated text will not be completely eliminated, preemptive strategies, e.g. better societal education on identifying fake reports, along with countermeasures, can help lessen the impact of disinformation attacks.\\

4. \textbf{Building alliances including civil society, international organisations and governments with technology providers, platforms and researchers for a coherent and proactive global strategy} - ecosystems built around AI technologies should be treated as complex systems and the necessity for a multidisciplinary approach to tackling the risks should be recognised (see e.g. \cite{joi_ito}).\\


The increasing convergence and ubiquity of AI technologies magnify the complexity of the challenges they present, and too often these complexities create a sense of detachment from their potentially negative implications. We must, however, ensure on a human level that these risks are assessed. Laws and regulations aimed at the AI space are urgently required and should be designed to limit the likelihood of those risks (and harms). With this in mind, the intent of this work is to raise awareness about the dangers of AI text generation to peace and political stability, and to suggest recommendations relevant to those in both the scientific and policy spheres that aim to address these challenges.

\section{Acknowledgements}

JB and MLO are with the United Nations Global Pulse innovation initiative supported by the Governments of Sweden, Netherlands and Germany and the William and Flora Hewlett Foundation. JB also is supported by the UK Science and Technology Facilities Council (STFC) grant number ST/P006744/1.

\bibliography{example_paper}

\begin{thebibliography}{20}
\providecommand{\natexlab}[1]{#1}
\providecommand{\url}[1]{\texttt{#1}}
\expandafter\ifx\csname urlstyle\endcsname\relax
  \providecommand{\doi}[1]{doi: #1}\else
  \providecommand{\doi}{doi: \begingroup \urlstyle{rm}\Url}\fi

\bibitem[Andersen et~al.(2018)]{accessnow_human_rights}
Andersen, L. et~al.
\newblock Human rights in the age of artificial intelligence.
\newblock \emph{Access Now}, 2018.

\bibitem[Baturo et~al.(2017)Baturo, Dasandi, and Mikhaylov]{UNGA_speeches}
Baturo, A., Dasandi, N., and Mikhaylov, S.~J.
\newblock Understanding state preferences with text as data: Introducing the un
  general debate corpus.
\newblock \emph{Research \& Politics}, 4\penalty0 (2), 2017.

\bibitem[Brundage et~al.(2018)]{OpenAI_review}
Brundage, M. et~al.
\newblock The malicious use of artificial intelligence: Forecasting,
  prevention, and mitigation.
\newblock 2018.
\newblock \textit{ar{X}iv preprint arXiv:1802.07228}.

\bibitem[{Council of Europe}(2017)]{algorithms_human_rights}
{Council of Europe}.
\newblock Algorithms and human rights: Study on the human rights dimension of
  automated data processing techniques and possible regulatory implications.
\newblock \emph{Council of Europe Study DGI(2017)12}, 2017.

\bibitem[Davidson et~al.(2017)Davidson, Warmsley, Marcy, and
  Weber]{hatespeech_offensive}
Davidson, T., Warmsley, D., Marcy, M., and Weber, I.
\newblock Automated hate speech detection and the problem of offensive
  language.
\newblock In \emph{Proceedings of the 11th International AAAI Conference on Web
  and Social Media}. {AAAI} Press, 2017.

\bibitem[Honnibal \& Montani(2017)Honnibal and Montani]{spacy2}
Honnibal, M. and Montani, I.
\newblock spacy 2: Natural language understanding with bloom embeddings,
  convolutional neural networks and incremental parsing.
\newblock \emph{To appear}, 2017.

\bibitem[Howard \& Ruder(2018)Howard and Ruder]{ULMFiT}
Howard, J. and Ruder, S.
\newblock Universal language model fine-tuning for text classification.
\newblock In \emph{Proceedings of the 56th Annual Meeting of the Association
  for Computational Linguistics (Volume 1: Long Papers)}, pp.\  328--339.
  Association for Computational Linguistics, 2018.

\bibitem[Howard et~al.(2018)]{fastai}
Howard, J. et~al.
\newblock fastai, 2018.

\bibitem[Ito(2018)]{joi_ito}
Ito, J.
\newblock \emph{The Practice of Change}.
\newblock PhD thesis, Keio Graduate School of Media and Governance, 2018.

\bibitem[Kshirsagar et~al.(2018)Kshirsagar, Cukuvac, McKeown, and
  McGregor]{hatespeech_embeddings}
Kshirsagar, R., Cukuvac, T., McKeown, K., and McGregor, S.
\newblock Predictive embeddings for hate speech detection on twitter.
\newblock In \emph{Proceedings of the 2nd Workshop on Abusive Language Online
  ({ALW}2)}, pp.\  26--32. Association for Computational Linguistics, 2018.

\bibitem[Merity et~al.(2017{\natexlab{a}})Merity, Keskar, and Socher]{AWD_LSTM}
Merity, S., Keskar, N.~S., and Socher, R.
\newblock Regularizing and optimizing {LSTM} language models.
\newblock 2017{\natexlab{a}}.
\newblock \textit{ar{X}iv preprint arXiv:1708.02182}.

\bibitem[Merity et~al.(2017{\natexlab{b}})]{wiki_pretrained}
Merity, S. et~al.
\newblock Pointer sentinel mixture models.
\newblock In \emph{Proceedings of the International Conference on Learning
  Representations (ICLR) 2017}, 2017{\natexlab{b}}.

\bibitem[{OpenAI}(2019)]{GPT-117M}
{OpenAI}.
\newblock {GPT-2}.
\newblock \url{https://github.com/openai/gpt-2}, 2019.

\bibitem[Radford et~al.(2018{\natexlab{a}})]{GPT2}
Radford, A. et~al.
\newblock Language models are unsupervised multitask learners.
\newblock \emph{OpenAI}, 2018{\natexlab{a}}.

\bibitem[Radford et~al.(2018{\natexlab{b}})]{OpenAI_blog}
Radford, A. et~al.
\newblock Better language models and their implications.
\newblock \emph{OpenAI}, 2018{\natexlab{b}}.

\bibitem[Risse(2018)]{carr_ai_human_rights}
Risse, M.
\newblock Human rights and artificial intelligence: An urgently needed agenda.
\newblock \emph{Carr Center for Human Rights Policy}, 2018.

\bibitem[Seymore \& Tull(2016)Seymore and Tull]{tweet_generation}
Seymore, J. and Tull, P.
\newblock Weaponizing data science for social engineering: Automated {E}2{E}
  spear phishing on twitter.
\newblock In \emph{Black Hat conference}, 2016.

\bibitem[Smith(2017)]{one_cycle}
Smith, L.~N.
\newblock Cyclical learning rates for training neural networks.
\newblock In \emph{Applications of Computer Vision (WACV), 2017 IEEE Winter
  Conference}, pp.\  464--472, 2017.

\bibitem[{Thies} et~al.(2016)]{face2face}
{Thies}, J. et~al.
\newblock Face2face: Real-time face capture and reenactment of rgb videos.
\newblock In \emph{2016 IEEE Conference on Computer Vision and Pattern
  Recognition (CVPR)}, pp.\  2387--2395, June 2016.

\bibitem[{UN General Assembly}(2018)]{unsr_human_rights}
{UN General Assembly}.
\newblock Report of the special rapporteur on the promotion and protection of
  the right to freedom of opinion and expression.
\newblock A/73/348, 2018.

\end{thebibliography}
\bibliographystyle{icml2019}

\end{document}